\documentclass[twocolumn]{article}
\usepackage{geometry}
\usepackage{times}      
\usepackage{mathptmx}   
\usepackage{helvet}     
\usepackage[numbers,sort&compress]{natbib}  
\usepackage{lipsum}     
\usepackage[numbers]{natbib}
\usepackage{formatting}
\usepackage{graphicx}
\usepackage{multirow}
\usepackage{xurl}
\usepackage[table,xcdraw]{xcolor}
\usepackage{afterpage}
\usepackage{flushend}
\usepackage{cuted}
\usepackage{caption}

\date{}
\begin{document}

\twocolumn[

\begin{@twocolumnfalse}

\title{From Generalist to Specialist: Improving Large Language \\
Models for Medical Physics Using ARCoT}

\author{
    Jace Grandinetti and
    Rafe McBeth\\\\
    \small{Department of Radiation Oncology, University of Pennsylvania}\\
    \small{\{jace.grandinetti, 
    rafe.mcbeth\}@pennmedicine.upenn.edu}
}

\maketitle

\begin{center}
    \begin{minipage}{0.9\textwidth}
        \begin{abstract}{
            \noindent
Large Language Models (LLMs) have achieved remarkable progress, yet their application in specialized fields, such as medical physics, remains challenging due to the need for domain-specific knowledge. This study introduces ARCoT (Adaptable Retrieval-based Chain of Thought), a framework designed to enhance the domain-specific accuracy of LLMs without requiring fine-tuning or extensive retraining. ARCoT integrates a retrieval mechanism to access relevant domain-specific information and employs step-back and chain-of-thought prompting techniques to guide the LLM’s reasoning process, ensuring more accurate and context-aware responses. Benchmarking on a medical physics multiple-choice exam, our model outperformed standard LLMs and reported average human performance, demonstrating improvements of up to 68\% and achieving a high score of 90\%. This method reduces hallucinations and increases domain-specific performance. The versatility and model-agnostic nature of ARCoT make it easily adaptable to various domains, showcasing its significant potential for enhancing the accuracy and reliability of LLMs in specialized fields.
            }
        \end{abstract}
    \end{minipage}
\end{center}

\bigskip

\end{@twocolumnfalse}
]

\section{Introduction}
Natural Language Processing (NLP) models have been experiencing a rapid rise in adoption across many domains due to the Transformer-based Large Language Models (LLMs) \citep{vaswani2017attention}. Many of the flagship and leading commercial LLMs have been trained across a vast dataset of knowledge to serve as a "generalist" artificial intelligence (AI) language model, allowing for use in a broad range of specialties. With a generalist approach, these models have performed remarkably well on a range of topics, spanning from English literature to computer programming. For instance, the leading state-of-the-art LLM model, GPT-4 (OpenAI, San Francisco, CA), scored $\sim 90^{th}$ percentile on the Bar Exam,  $\sim99^{th}$ percentile on the GRE Verbal, and a $75$\% on the Medical Knowledge Self-Assessment Program \citep{achiam2023gpt}.

\afterpage{
\begin{figure*}[ht]
    \centering
    \includegraphics[width=1.8\columnwidth]{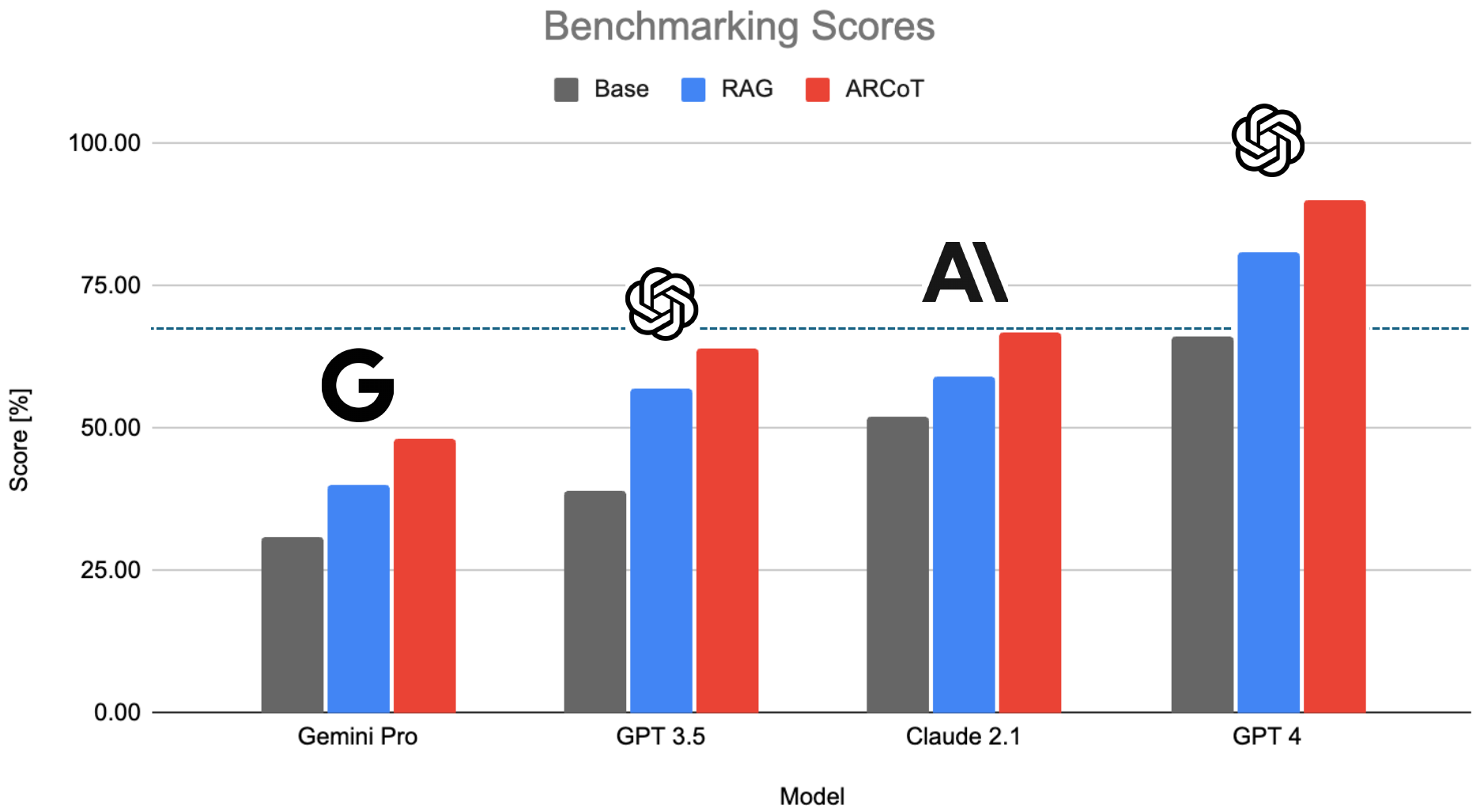}
    \caption{Bar graph illustrating the scores of four leading model configurations on a medical physics benchmark exam, highlighting the enhanced performance of our proposed ARCoT model that achieved a high score of up to 90\%. The dashed line represents an average human score of 68\% achieved by a cohort of medical physicists on a similar exam, serving as a comparative human benchmark.}
    \label{fig:bar_chart}
\end{figure*}
}

However, despite these advancements, a disparity in knowledge for highly specialized domains (e.g., medical physics, radiation oncology, surgery) has been observed due to limited datasets used when training these large models \citep{beaulieu2024evaluating, liu2023tailoring, van2023exploration}. 
A study evaluating the performance of GPT-3.5 and GPT-4 on a medical physics exam found that the base models scored only $35$\% and $67$\% respectively \citep{holmes2023evaluating}, while GPT-4 similarly scored $68$\% on a surgical knowledge test \citep{beaulieu2024evaluating}. 
LLMs also lack access to updated information due to a knowledge cutoff date and lack the ability to reliably source citations from the trained data. Moreover, these models often experience hallucinations \citep{lee2018hallucinations, bang2023multitask, rawte2023survey}, outputting coherent but incorrect information. In the domain of clinical implementation, falsified or misleading outputs of LLM models can be especially dangerous as healthcare professionals may use the outputs to guide their decisions, ultimately impacting patient care.

To address these concerns, four major solutions exist with varying degrees of complexity: (1) training a new language model from scratch, (2) fine-tuning pre-trained models, (3) Retrieval Augmented Generation (RAG) and (4) prompt engineering. The most complex approach is to develop a new large language model from scratch, which would give users the most control over the information used for training. While this might seem like the most straightforward solution, developing new LLMs can be costly, with state-of-the-art LLMs requiring \$10 to \$100 million in resources and infrastructure \citep{umeton2024gpt, peng2023study}. These kind of resources may be prohibitive for many hospitals and clinics to implement, necessitating alternative approaches for bringing LLMs into the workflow.

Fine-tuning pre-trained LLMs involves a secondary training step that incorporates smaller and more specific data to be used to adjust a subset of the weights, with the goal of producing a model that has improved performance on specific tasks. While development of fine-tuning techniques efforts are ongoing, robust solutions remain an open challenge \citep{zheng2024fine, radiya2020fine, dodge2020fine, ziegler2019fine, ding2023parameter, chen2023longlora}. Fine-tuning has been shown to be resource-intensive, requiring highly curated datasets with trial-and-error approaches to guide the model to the desired outputs. More importantly, it's also been demonstrated that fine-tuning methods can still lead to undesired outputs, loss of prior knowledge, and overall degraded performance \citep{luo2023empirical, goodfellow2013empirical}.

As an alternative to training and fine-tuning, RAG has rapidly emerged as a robust method for enhancing LLM models in domain-specific areas without the need for additional training \citep{AI}. This is accomplished by employing a retriever to locate user-stored information corresponding to the user’s inquiry, supplying this context to the transformer model during the inference process. This has been shown to significantly improve output accuracy and reduce hallucinations by providing the necessary contextual information to augment the model's knowledge base without the need for fine-tuning or training new models from scratch \citep{wu2023ragtruth, tonmoy2024comprehensive}.
Although RAG has demonstrated model improvements, its current limitations, specifically in handling context length (i.e., number of tokens) and effectively accessing relevant information from retrieved contexts, have posed significant challenges. It has been observed that LLMs struggle with managing information when dealing with lengthy input contexts, which restricts their ability to handle large databases or extensive documents effectively \citep{liu2024lost}. One emerging method to help address this limitation is to use re-ranking transformers which can help to improve similarity of retrieved content, thereby reducing the total amount of context required to input into the model (see Sec.~\ref{rerank}). 

Prompt engineering has become a ubiquitous technique employed in LLMs to enhance model outputs. By providing initial instructions to the model, users can guide the model during inference and maximize its potential, resulting in higher quality answers \citep{chen2023unleashing, marvin2023prompt}. One prominent method for this is known as Chain of Thought (CoT) prompting, which instructs the model to break down complex problems into smaller, incrementally solvable sub-problems. Extensive research has demonstrated that CoT prompting significantly improves model performance, particularly for queries that involve multi-step reasoning and calculations \citep{wei2022chain, hongru2023cue, wang2023chain}.

In this study, we propose a framework that we named \textbf{A}daptable \textbf{R}etrieval-based \textbf{C}hain \textbf{o}f \textbf{T}hought (ARCoT) that utilizes RAG and CoT prompting techniques to improve the performance of LLM models within the domain-specific field of radiation oncology medical physics. We benchmarked this model using a multiple-choice medical physics exam to compare performance against base LLM models and a reported average score of a human cohort of medical physicists who participated in a similar exam \citep{holmes2023evaluating}. This technique is easily adaptable to different domains using various LLM models, highlighting the robust potential of this approach.

\section{Methods}

The proposed framework uniquely combines RAG with advanced prompting techniques to enhance overall model performance. This integration includes a Step-Back (SB) prompting strategy, which optimizes the relevance of retrieved documents. These documents are then refined by a re-ranking transformer that prioritizes selections most relevant to the input query. Additionally, Chain-of-Thought (CoT) prompting is incorporated to further enhance reasoning capabilities during inference. Detailed explanations of these individual methods are provided in Sections \ref{RAG} - \ref{arcot}.

\begin{figure*}[htbp]
    \centering
    \includegraphics[width=\textwidth]{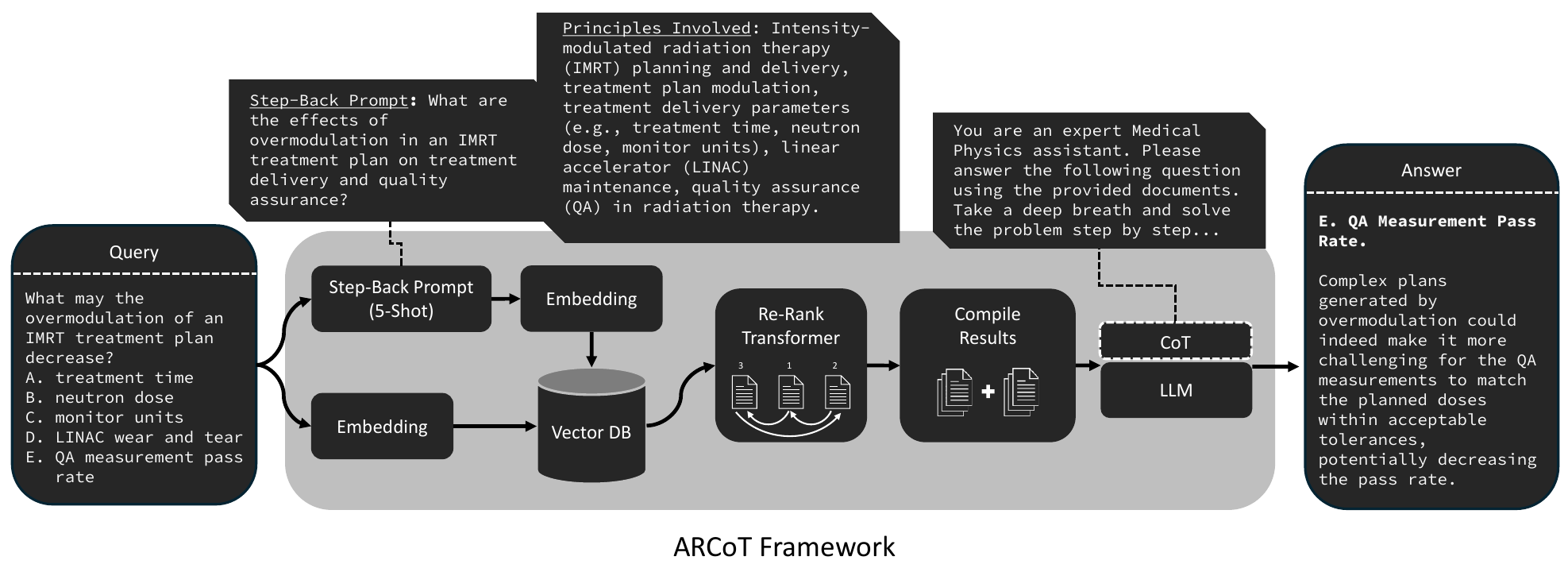}
    \caption{Architecture of the proposed ARCoT framework with an example of a user query. A hybrid SB prompting approach is implemented after the input query to improve similarity results with the original embedded prompt. A re-ranking transformer filters results with the highest relevance and a CoT prompt is used to further enhance model inference.}
    \label{fig:flow_chart}
\end{figure*}

\subsection{Retrieval Augmented Generation} \label{RAG}
A vector database was created by aggregating open-source reports (e.g., AAPM Task Groups and MPPGs), scholarly publications, and textbooks (e.g., IAEA Radiation Oncology Physics) relevant to medical physics. Our objective was to compile a dataset that extensively covered a wide array of fundamental topics and their associated practices within medical physics with a focus on prominent areas like quality assurance, treatment planning, radiation safety, and medical imaging. 
A collection of around 60 documents was compiled and converted into text files by employing an open-source Python library \citep{Unstructured-IO}. The parsed documents were then dynamically chunked using a recursive approach that aims to minimize chunk size variation without separating relevant information. This is accomplished with a hierarchical and iterative approach that uses stop separators and can recursively adjust the chunk sizes if they vary too much. The maximum chunk size was set to $500$, with a chunk overlap of $50$. This resulted in a vector database sized at roughly $10,000$ vectors. Each chunk was semantically embedded using OpenAI's embedding model (text-embedding-ada-002). The embeddings were stored in a vector database (Pinecone, San Francisco, CA) where cosine similarity was used to retrieve vector embeddings with the highest similarity indexes.

\subsection{Step-Back Prompting} \label{SB}
Prompt engineering has demonstrated success at improving model inferences by adding a layer of instructions that can help guide the LLMs as they make inferences. This has been notably demonstrated in Chain-of-Thought (CoT) prompting where multi-step reasoning is employed to solve intermediate steps from complex queries \citep{wei2022chain, zhang2023multimodal}. A unique application of this CoT prompting is a method proposed by Google's DeepMind known as Step-Back (SB) prompting \citep{zheng2023take} which creates a secondary query and attempts to break down and simplify a question that has multiple components. Expanding on this work, we found that we were able to apply this prompting strategy to generate a second query that is more fundamental (i.e. has taken a "step-back" from the initial query) and then generate important keywords that might be specific to the topic. This inference was designed to output a step-back prompt and involved key principles (see Fig.~\ref{fig:flow_chart}). This inference was also further improved with 5-shot prompting, which provided the model with additional context and examples to enhance its ability to generate more accurate and contextually relevant responses.

The main reasoning of this modification is that in highly specialized fields where the LLMs have not been sufficiently trained, reliance on the vector similarity search is critical so the LLM has the appropriate context to answer the question. Queries in specialized domains may be missed in the similarity search since specific keywords may not be included in the original query, and the embedding models may struggle to capture the semantic meaning of the query accurately. In order to preserve key information and associated nuances that could be important to answering the question, the original query is also passed, resulting in a hybrid or blended RAG approach \citep{sawarkar2024blended}.

\subsection{Re-Ranking Transformer} \label{rerank}
Language models have been shown to encounter degraded performance when dealing with long contexts, a problem known as "lost in the middle" where the LLM will use more relevant information at the beginning (primacy bias) or at the end (recency bias) \citep{liu2024lost}. In these situations, the models struggle to maintain contextual information from retrieved documents, leading to ineffective query answering. To mitigate this issue and optimize the number of retrieved documents required for answering queries, we employed a re-ranking transformer endpoint (Cohere, Toronto, CA) designed specifically for contextually compressing search results \citep{gilbert2023semantic}. This secondary transformer was used to filter and retain only the top re-ranked documents, minimizing the input context length and reducing overall token usage. This approach can be especially useful for API-dependent models, making it a cost-effective solution. We varied the number of retained documents, each containing approximately 500 tokens, from $2$ to $25$. Our findings demonstrated diminishing returns beyond 8 documents, considering the trade-off between token usage and speed. Moreover, providing the model with more than roughly $20$ documents as context often resulted in some missed information, consistent with the "lost in the middle" trend observed in RAG retrieval. To optimize a balance between providing sufficient context and the constraints imposed by model token limits and reduced performance with lengthy contextual inputs, we retrieved a total of 50 top similarity results (25 from each query) and selected the top 8 highest similarity documents after the re-ranking process.

\subsection{ARCoT Framework} \label{arcot}
When constructing this LLM framework, our goal was to create an adaptable framework that could be deployed on existing state-of-the-art LLMs without the need for fine-tuning or retraining. The framework we propose only uses the pre-trained LLM for inferences, boosting performance primarily through RAG retrieval. A last step for further enhancing the model's performance is CoT prompting which asks the model to think carefully step-by-step. It's important to note that we prompted the model to output its reasoning as it solved each problem to minimize guessing and to evaluate where the model may be making mistakes (i.e. retrieval step, incorrect calculation, hallucination, etc.). Fig.~\ref{fig:flow_chart} illustrates the proposed model framework where the input query is split for embedding and SB prompting. The top $25$ similarity results are retrieved for each route which are then filtered down to a total of 8 through the re-ranking transformer before combining and passing through the final CoT inference.

\subsection{Benchmarking Exam} \label{benchmark}
To benchmark LLMs in a way that reduces subjective bias, multiple-choice questions are often used for evaluations. Since there are no existing benchmarking exams in medical physics, we compiled a set of 128 multiple-choice questions, each with 4-5 possible answers, from the RAPHEX 2023 Therapy exam. This was done to guarantee a fair and broad representation of the subject matter, recognizing that crafting a new exam from scratch could unintentionally introduce bias. We omitted questions that couldn't be answered without reference to tables or images, as our model was not multi-modal and able to process visual information. A prior study from \citeauthor{holmes2023evaluating} reported that their selected group of medical physicists achieved an average score of $68$\% on a set of similar RAPHEX questions, many of which were identical and used in our study. We present these findings to establish a baseline for an approximation on expected human performance.

One drawback to benchmarking with multiple-choice questions is that the method does not inherently penalize correct guesses. Given the potential for using LLMs in clinical settings, it's imperative that the model demonstrates a reliable understanding of the correct answers and is not hallucinating or generating them by chance. To mitigate this issue, we subjected each question to the model five separate times. The model had to answer each question correctly in all five attempts, otherwise no points were awarded. This strictly penalizes any incorrect answers and reduces the probability of scoring points through random guessing to less than $0.1$\%.

\begin{table*}[htbp]
\centering
\resizebox{\textwidth}{!}{%
\begin{tabular}{clcccccccc}
\hline
\multicolumn{2}{c}{\textbf{LLM Model}} &
  \multicolumn{1}{c}{\textbf{\begin{tabular}[c]{@{}c@{}}Physics\\ Fundamentals\end{tabular}}} &
  \multicolumn{1}{c}{\textbf{\begin{tabular}[c]{@{}c@{}}Therapy\\ Fundamentals\end{tabular}}} &
  \multicolumn{1}{c}{\textbf{\begin{tabular}[c]{@{}c@{}}Treatment\\ Planning\end{tabular}}} &
  \multicolumn{1}{c}{\textbf{\begin{tabular}[c]{@{}c@{}}Radiation\\ Safety\end{tabular}}} &
  \multicolumn{1}{c}{\textbf{Imaging}} &
  \multicolumn{1}{c}{\textbf{\begin{tabular}[c]{@{}c@{}}Advanced\\ Treatments\end{tabular}}} &
  \multicolumn{1}{c}{\textbf{\begin{tabular}[c]{@{}c@{}}Calculation\\ Based\end{tabular}}} &
  \multicolumn{1}{c}{\textbf{All}} \\ \hline
 &
  \textbf{Base} &
  \textbf{85.71\%} &
  36.67\% &
  20.00\% &
  30.00\% &
  41.67\% &
  38.71\% &
  6.67\% &
  39.06\% \\
 &
  \textbf{RAG} &
  \textbf{85.71\%} &
  \textbf{66.67\%} &
  40.00\% &
  54.55\% &
  50.00\% &
  61.29\% &
  40.00\% &
  57.03\% \\
\multirow{-3}{*}{\textbf{GPT-3.5}} &
  \textbf{ARCoT} &
  \textbf{85.71\%} &
  \textbf{66.67\%} &
  {\color[HTML]{000000} \textbf{53.33\%}} &
  \textbf{63.64\%} &
  \textbf{75.00\%} &
  \textbf{64.52\%} &
  \textbf{46.67\%} &
  {\color[HTML]{000000} \textbf{65.63\%}} \\ \hline
 &
  \textbf{Base} &
  \textbf{100.00\%} &
  70.00\% &
  46.67\% &
  36.36\% &
  75.00\% &
  74.19\% &
  53.33\% &
  {\color[HTML]{000000} 66.41\%} \\
 &
  \textbf{RAG} &
  \textbf{100.00\%} &
  \textbf{83.33\%} &
  56.67\% &
  81.82\% &
  83.33\% &
  90.32\% &
  60.00\% &
  81.25\% \\
\multirow{-3}{*}{\textbf{GPT-4}} &
  \textbf{ARCoT} &
  \textbf{100.00\%} &
  \textbf{83.33\%} &
  \textbf{80.00\%} &
  \textbf{90.91\%} &
  \textbf{83.33\%} &
  \textbf{96.77\%} &
  \textbf{86.67\%} &
  \textbf{89.84\%} \\ \hline
 &
  \textbf{Base} &
  \textbf{92.86\%} &
  43.33\% &
  43.33\% &
  18.18\% &
  58.33\% &
  58.06\% &
  13.33\% &
  51.56\% \\
 &
  \textbf{RAG} &
  71.43\% &
  60.00\% &
  36.67\% &
  \textbf{72.73\%} &
  \textbf{75.00\%} &
  61.29\% &
  33.33\% &
  58.59\% \\
\multirow{-3}{*}{\textbf{Claude 2.1}} &
  \textbf{ARCoT} &
  85.71\% &
  \textbf{80.00\%} &
  \textbf{50.00\%} &
  54.55\% &
  \textbf{75.00\%} &
  \textbf{64.52\%} &
  \textbf{53.33\%} &
  \textbf{67.19\%} \\ \hline
 &
  \textbf{Base} &
  50.00\% &
  26.67\% &
  16.67\% &
  27.27\% &
  \textbf{75.00\%} &
  25.81\% &
  6.67\% &
  31.25\% \\
 &
  \textbf{RAG} &
  64.29\% &
  46.67\% &
  10.00\% &
  45.45\% &
  66.67\% &
  \textbf{38.71\%} &
  13.33\% &
  39.84\% \\
\multirow{-3}{*}{\textbf{Gemini Pro}} &
  \textbf{ARCoT} &
  \textbf{78.57\%} &
  \textbf{56.67\%} &
  \textbf{23.33\%} &
  \textbf{63.64\%} &
  66.67\% &
  \textbf{38.71\%} &
  \textbf{26.67\%} &
  \textbf{48.44\%} \\ \hline
\end{tabular}%
}
\caption{Percentage of correct responses across seven categories of the 2023 RAPHEX Therapy multiple-choice exam. The models enhanced with our proposed ARCoT framework scored highest or outperformed the baseline in all but 3 benchmarks, with notable advancements in Treatment Planning and Calculation-Based questions.}
\label{tab:results}
\end{table*}

\begin{figure*}[htbp]
    \centering
    \includegraphics[width=2\columnwidth]{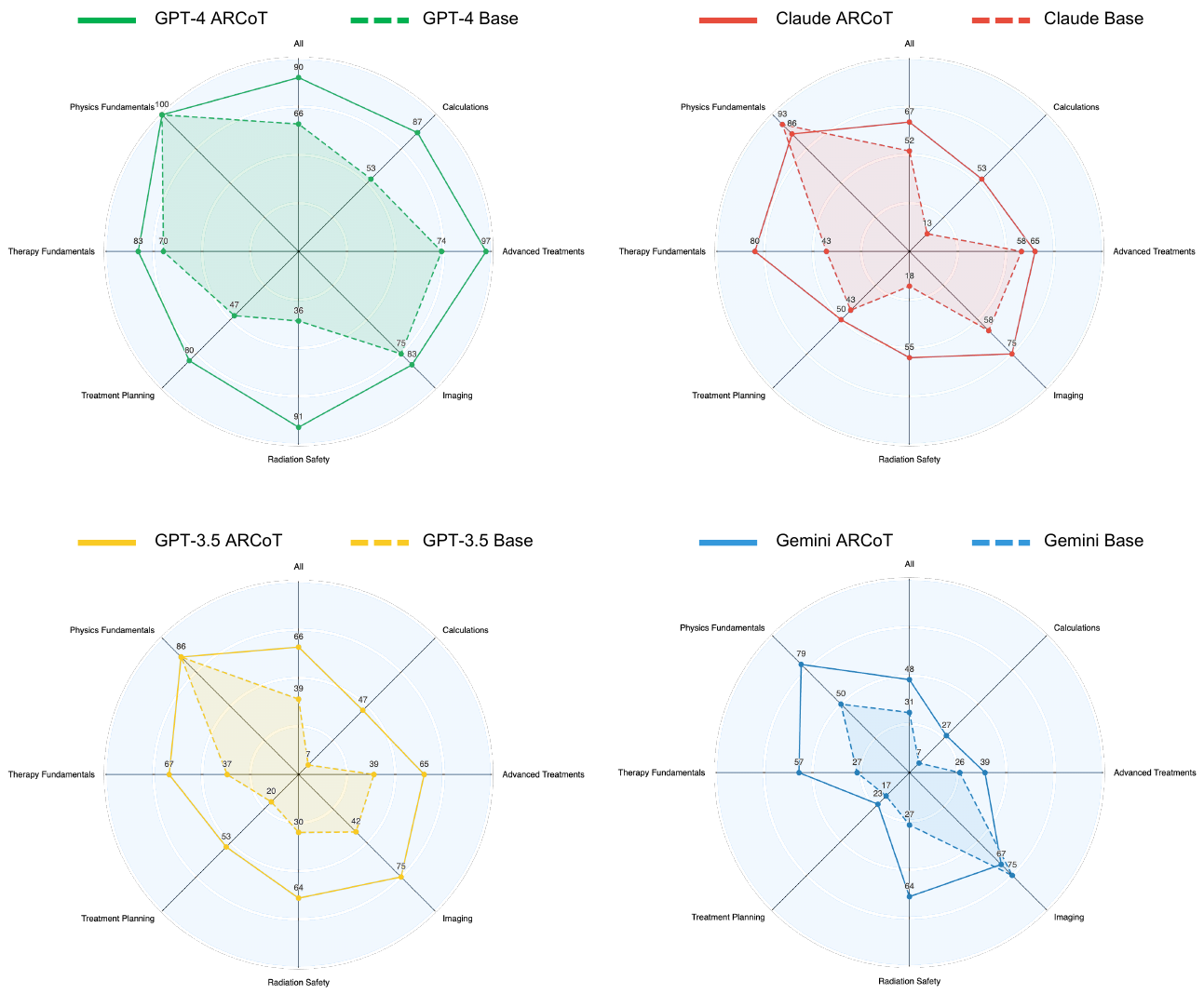}
    \caption{Radar plots depicting the benchmark scores of each LLM using the ARCoT framework (solid) against the base model (dashed and filled). Edges of each plot correspond to a score of 100\%.}
    \label{fig:radar}
\end{figure*}

\section{Results}
A total of 4 leading state-of-the-art models were benchmarked: Gemini Pro 1.0 (Google), GPT-3.5 and GPT-4 (OpenAI), and Claude 2.1 (Anthropic). Each model was benchmarked with and without RAG to serve as a control against our proposed ARCoT framework. The exam's questions were sorted into six categories, with calculation-based questions being doubly classified under their respective topic and the calculation category. The data from Table~\ref{tab:results} and Figs. \ref{fig:bar_chart} (summarized) and \ref{fig:radar} (categorized) show the final results of our model on the benchmark exam. Our ARCoT framework was shown to enhance performance with an average improvement of $47\%$ compared to the base model and $15\%$ improvement compared to RAG alone. Our framework also ranked highest in all sections or outperformed the others in $25$ of the $28$ model categories.
Notably, GPT-3.5 experienced the largest improvement of 68\% while GPT-4's performance increased from 67\% (just below the previously reported human performance of 68\%) to 90\%.

\section{Discussion}
The largest improvements were observed with GPT-3.5, likely due to its limitations on training data compared to GPT-4, Claude, and Gemini Pro. Calculation improvements were also notable, largely due to the CoT prompting and enhanced recall of relevant equations. Furthermore, we observed that a more advanced model like GPT-4 could retrieve pertinent information and process in-context learning with greater efficiency, even though GPT-3.5 showed larger improvements overall. This suggests that integrating RAG and advanced prompting with more capable models has a potential synergistic effect, as the foundational knowledge is utilized effectively during the information retrieval and inference processes. Claude 2.1, allegedly developed with AI safety considerations, exhibited performance comparable to GPT-4 in the baseline evaluation. However, when benchmarked using the ARCoT and RAG methods, Claude 2.1 underperformed in two out of seven categories. We hypothesize that this discrepancy can be attributed to the training and tuning methodology employed in the development of Claude 2.1. The model appears to restrict query inferences based solely on the provided retrieved context. While we speculate that this approach is intended to mitigate hallucinations, it may have inadvertently decreased performance by constraining answers to incomplete information that was insufficient to comprehensively address the query. These findings indicate the potential for further refinement of this method through fine-tuning and retraining techniques to optimize model behavior, rather than relying exclusively on inherent knowledge.

This study has several limitations, including the use of a non-comprehensive dataset due to restrictions on open-source content and information. We anticipate that expanding our dataset to encompass a broader range of medical physics knowledge could lead to improved results. Furthermore, implementing data pre-processing steps to clean the dataset has the potential to enhance precision. It is important to note that this study benchmarked the performance of state-of-the-art commercial LLM models. Future research would benefit from a comparative analysis using a more diverse set of models, varying in size, and training sets, and incorporating open-source options, to gain a more comprehensive understanding of performance across different architectures and training methodologies.

\section{Conclusion}
In conclusion, we have proposed ARCoT: an Adaptable Retrieval Augmented Chain of Thought framework that can be easily implemented and deployed on existing LLM models to improve performance in highly specialized domains. From a limited database of open-source materials, we were able to improve the benchmarking performance of the leading GPT-4 model from a score of $67$\% to $90$\% in the field of medical physics without the need for fine-tuning or building a new LLM model from scratch. This highlights the technique's robust potential for easy adaptation to various domains, while remaining independent of specific LLM models, demonstrating its applicability beyond medical physics.

\bibliographystyle{unsrtnat}
\bibliography{main}

\end{document}